\def\year{2019}\relax

\documentclass[letterpaper]{article} 
\usepackage{aaai19}  
\usepackage{times}  
\usepackage{helvet}  
\usepackage{courier}  
\usepackage[hyphens]{url} 
\usepackage{graphicx}  
\usepackage{graphicx} 
\usepackage{amsmath}
\usepackage{enumitem}

\title{Unsupervised Cross-spectral Stereo Matching by Learning to Synthesize} 

\author{Mingyang~Liang$^{1,2}$\footnotemark[1], Xiaoyang~Guo$^{3}$\thanks{These authors contributed equally to this work.}, Hongsheng~Li{$^3$}, Xiaogang~Wang{$^{3}$}, You~Song$^{1}$ \thanks{Corresponding author.} \\
{$^1$Beihang University, Beijing, China}\\
{$^2$SenseTime Research}\\
{$^3$The Chinese University of Hong Kong, Hong Kong, China}\\
{\{liangmingyang,songyou\}@buaa.edu.cn, \{xyguo, hsli, xgwang\}@ee.cuhk.edu.hk}
}

\begin{document}
\maketitle
\begin{abstract}
\noindent Unsupervised cross-spectral stereo matching aims at recovering disparity given cross-spectral image pairs without any depth or disparity supervision. The estimated depth provides additional information complementary to original images, which can be helpful for other vision tasks such as tracking, recognition and detection.
However, there are large appearance variations between images from different spectral bands, which is a challenge for cross-spectral stereo matching. Existing deep unsupervised stereo matching methods are sensitive to the appearance variations and do not perform well on cross-spectral data.
We propose a novel unsupervised cross-spectral stereo matching framework based on image-to-image translation. First, a style adaptation network transforms images across different spectral bands by cycle consistency and adversarial learning, during which appearance variations are minimized. Then, a stereo matching network is trained with image pairs from the same spectra using view reconstruction loss. At last, the estimated disparity is utilized to supervise the spectral translation network in an end-to-end way. Moreover, a novel style adaptation network F-cycleGAN is proposed to improve the robustness of spectral translation.
Our method can tackle appearance variations and enhance the robustness of unsupervised cross-spectral stereo matching. Experimental results show that our method achieves good performance without using depth supervision or explicit semantic information.

\end{abstract}

\section{Introduction}

\noindent Multi-camera multi-spectral systems have become very common in many modern devices like Realsense, Kinect, and iPhoneX. Moreover, it has been proven that infrared images are very helpful in face recognition~\cite{lezama2017not}, detection~\cite{xu2017learning}, and scene parsing~\cite{st2017mutual}.

\begin{figure}[!ht]
    \centering
    \begin{tabular}{@{}c@{}c}
        \begin{tabular}{@{}c}
        \includegraphics[width=0.460\linewidth]{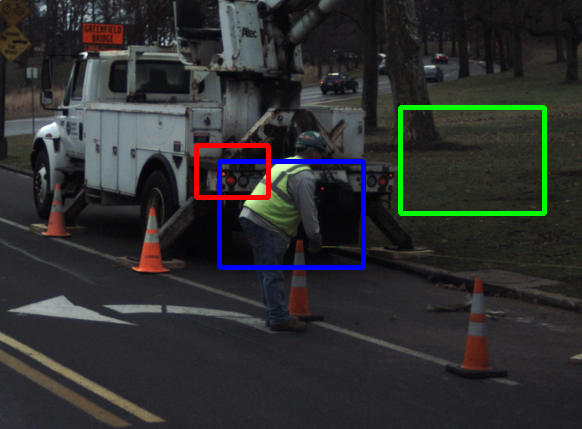}
        \end{tabular}
        &
        \begin{tabular}{@{}c}
        \includegraphics[width=0.460\linewidth]{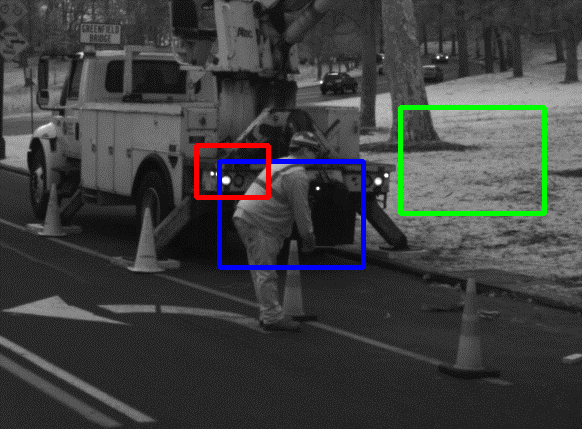}
        \end{tabular}
        \\
        (a) Left VIS & (b) Right NIR
        \\
        \begin{tabular}{@{}c}
        \includegraphics[width=0.460\linewidth]{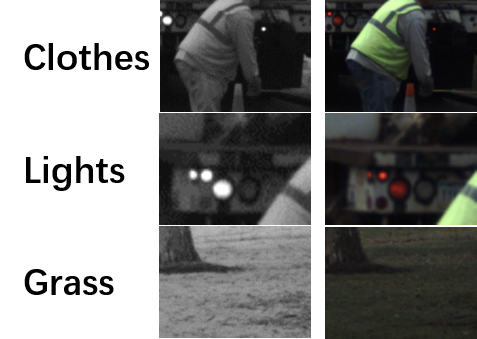}
        \end{tabular}
        &
        \begin{tabular}{@{}c}
        \includegraphics[width=0.460\linewidth]{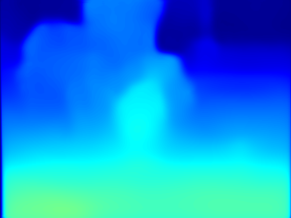}
        \end{tabular}
        \\
        (c) Appearance variations & (d) Predicted disparity 
    \end{tabular}
    \caption{The appearance variations between cross-spectral image pairs. VIS and NIR represent visible and near-infrared images respectively. Almost invisible light sources in the VIS turn into a dazzling white in the NIR. Grassland also shows huge illumination difference between VIS and NIR. The goal of cross-spectral stereo matching is to overcome the appearance differences between different spectra and predict accurate disparity.}
    \label{fig:contrast}
\end{figure}

\begin{figure*}[t]
\centering
\includegraphics[width=\linewidth]{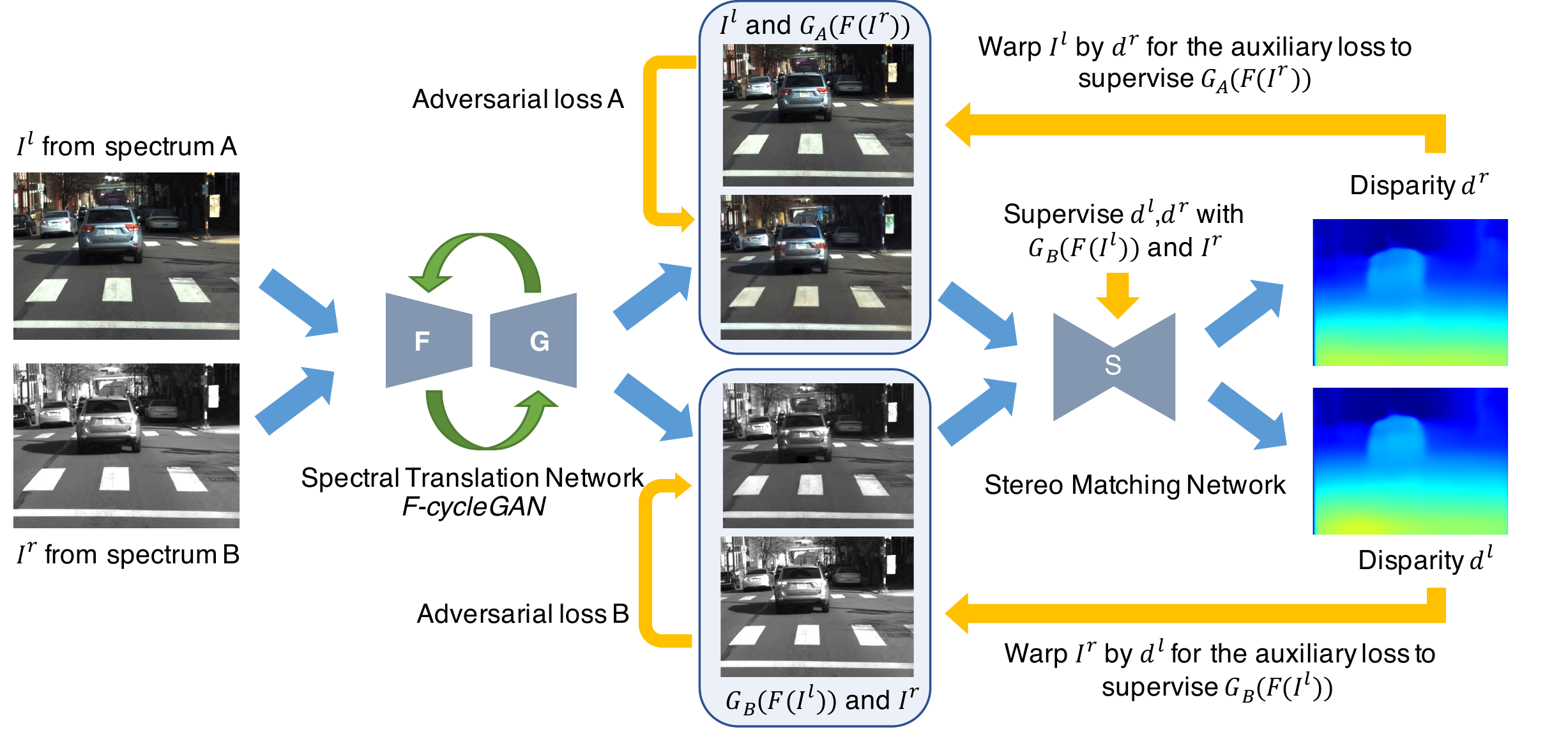}
\caption{The structure of our proposed cross-spectral stereo matching framework. First, we input $\{I^l, I^r\}$ into the spectral translation network (STN) to get $G_B(F(I^l))$ and $G_A(F(l^r))$. The \textbf{F} network maps the images into hidden feature space $X$ and \textbf{G} networks ($G_A$ and $G_B$) convert feature $X$ back into images of spectrum A and B. The stereo matching network (SMN) takes $I^{l,r}$, $G_B(F(I^l))$ and $G_A(F(l^r))$ to predict disparities $\{d^l, d^r\}$. By well-designed loss functions, we can train the STN and SMN jointly without depth ground-truth, extra-data or human-intervention. }
\label{fig:framework}
\end{figure*}

Stereo matching is one of the most heavily investigated topics in computer vision \cite{scharstein2002taxonomy}. Given a rectified image pair ($I_l$ for the left image, $I_r$ for the right image), stereo matching focuses on finding correspondence of each pixel between two images. If the right pixel $I_r(x{-}d, y)$ corresponds to the left pixel $I_l(x, y)$, then we can define $d$ as the disparity of the pixel $I_l(x, y)$. Moreover, if we know the camera's focal length $f$ and the distance $B$ between the two camera centers, the disparity can be converted into depth by $fB/d$.

Cross-spectral stereo matching is stereo matching for images from different spectra, for example the left image is a visible image and the right image is a near-infrared image in Fig.~\ref{fig:contrast}. The recovered depth provides additional information which is complementary to semantic features of individual spectrum. In addition, the estimated depth can help improve missing areas of depth images captured by depth sensors (e.g. reflection or transparent surfaces) \cite{chiu2011improving}.  

However, the cross-spectral stereo matching is still a challenging task especially without depth supervision \cite{zhi2018deep}, because there are great illumination differences between images of different spectra. The translation between different spectra is quite complex and hard to accurately describe with a simple linear transformation. Figure \ref{fig:contrast} shows an example of cross-spectral (visible and near-infrared) stereo matching.

The key of traditional cross-spectral stereo matching is to design robust descriptors or features between the two modalities, such as ANCC~\cite{heo2011robust} and DASC~\cite{kim2015dasc}. However, these traditional methods are still not robust enough for transparent objects and large illumination variations. Zhi et al.~\cite{zhi2018deep} proposed deep material-aware cross-spectral stereo matching, which tried to tackle the problem with deep neural networks and unsupervised learning. However, this method suffers from severe limitations: (i) The method requires additional semantic annotations to obtain auxiliary material information. (ii) The loss function is manually designed for different materials, which limits its applications to other scenarios.

To tackle the above problems, in this paper we employ image-to-image translation to assist cross-spectral stereo matching, and our full framework is shown in Figure \ref{fig:framework}. By regarding the difference between different spectral images as the different distributions, we explore the possibility of applying image-to-image translation methods to assist unsupervised cross-spectral stereo matching. We use two networks to transform images across different spectral bands and estimate disparity respectively. The first network is a spectral translation network (STN), which transforms images by cycle consistency and adversarial learning. The second network is a stereo matching network (SMN), which is trained with the image pairs transformed to the same spectrum by the spectral translation network. Then, we use the disparity predicted by the SMN to supervise the spectral translation network again. A novel share-encoder spectral translation network F-cycleGAN is employed to make the whole framework more robust.

Our contributions are as follows:

\begin{itemize}[noitemsep,nolistsep]

\item We proposed a novel framework for cross-spectral stereo matching, which iteratively optimizes spectral translation network and stereo matching network.

\item The proposed F-cycleGAN based on image-to-image translation and adversarial learning improves the robustness of image transformation.

\item Our method surpasses state-of-the-art methods on cross-spectral stereo matching without depth supervision and extra human intervention.

\end{itemize}
\begin{figure*}[!t]
    \centering
    \begin{tabular}{@{}c@{}c@{}c@{}c@{}c@{}c@{}c}
        \begin{tabular}{@{}c}
        \includegraphics[width=0.125\linewidth]{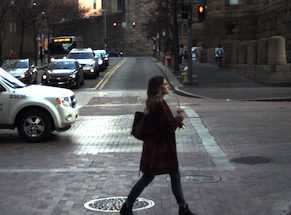}
        \end{tabular}
        &
        \begin{tabular}{@{}c}
        \includegraphics[width=0.125\linewidth]{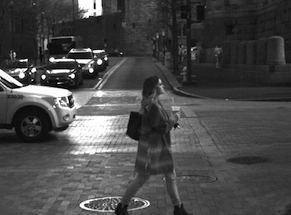}
        \end{tabular}
        &
        \begin{tabular}{@{}c}
        \includegraphics[width=0.125\linewidth]{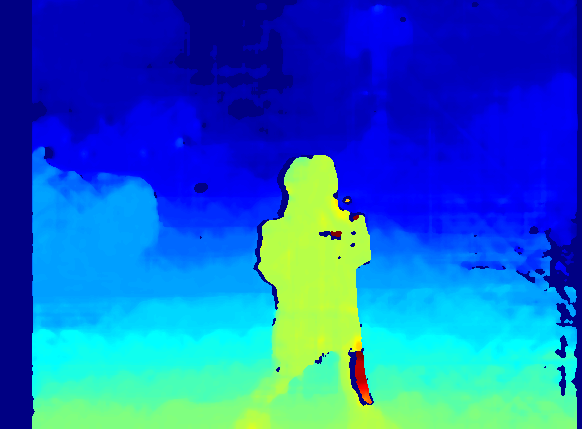}
        \end{tabular}
        &
        \begin{tabular}{@{}c}
        \includegraphics[width=0.125\linewidth]{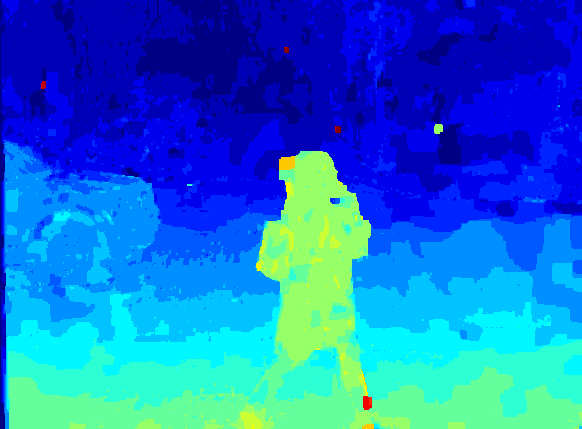}
        \end{tabular}
        &
        \begin{tabular}{@{}c}
        \includegraphics[width=0.125\linewidth]{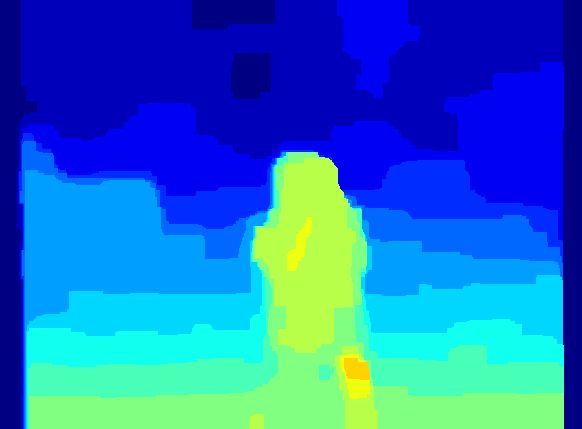}
        \end{tabular}
        &
        \begin{tabular}{@{}c}
        \includegraphics[width=0.125\linewidth]{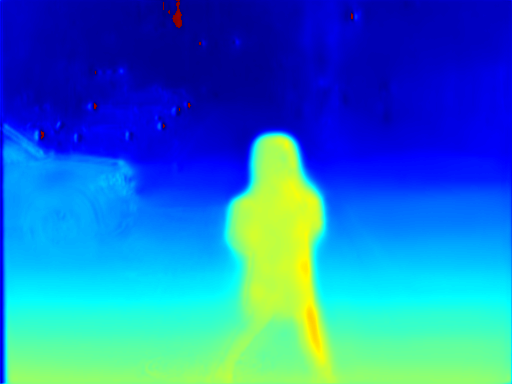}
        \end{tabular}
        &
        \begin{tabular}{@{}c}
        \includegraphics[width=0.125\linewidth]{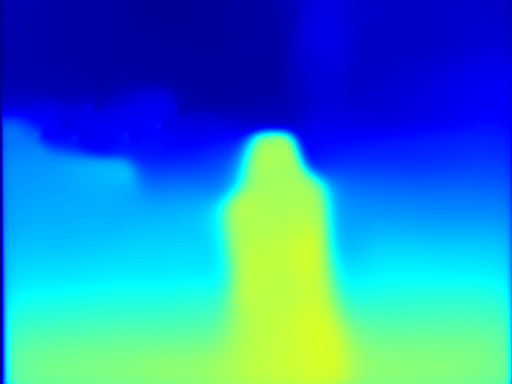}
        \end{tabular}
        \\
        \begin{tabular}{@{}c}
        \includegraphics[width=0.125\linewidth]{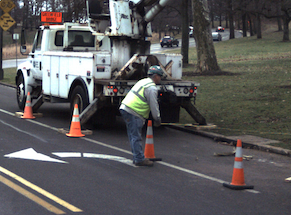}
        \end{tabular}
        &
        \begin{tabular}{@{}c}
        \includegraphics[width=0.125\linewidth]{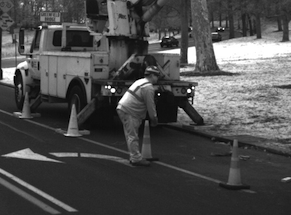}
        \end{tabular}
        &
        \begin{tabular}{@{}c}
        \includegraphics[width=0.125\linewidth]{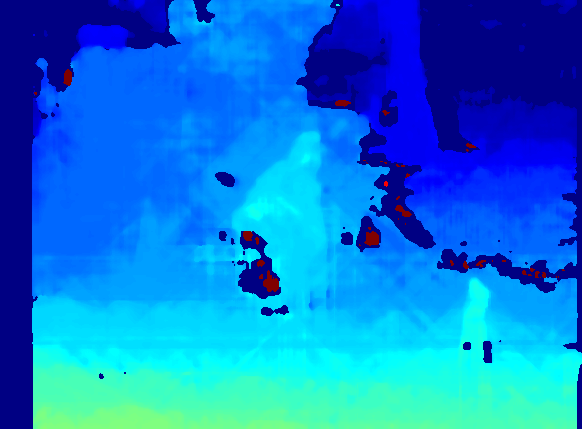}
        \end{tabular}
        &
        \begin{tabular}{@{}c}
        \includegraphics[width=0.125\linewidth]{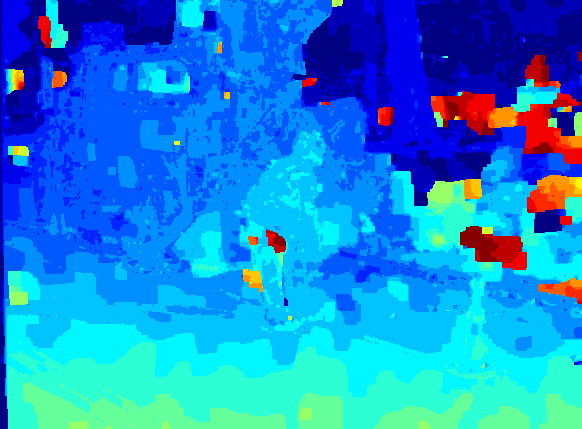}
        \end{tabular}
        &
        \begin{tabular}{@{}c}
        \includegraphics[width=0.125\linewidth]{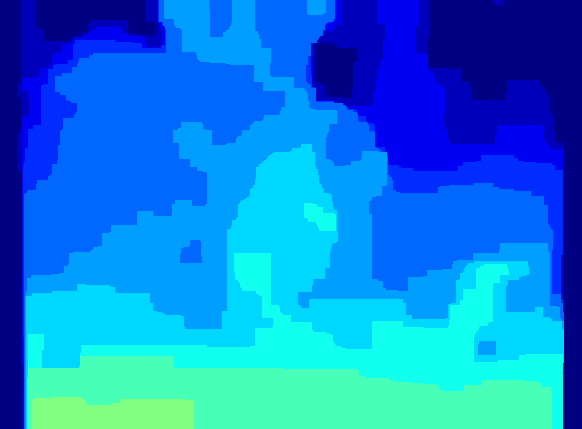}
        \end{tabular}
        &
        \begin{tabular}{@{}c}
        \includegraphics[width=0.125\linewidth]{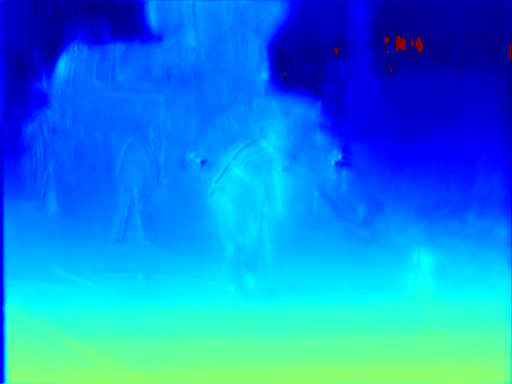}
        \end{tabular}
        &
        \begin{tabular}{@{}c}
        \includegraphics[width=0.125\linewidth]{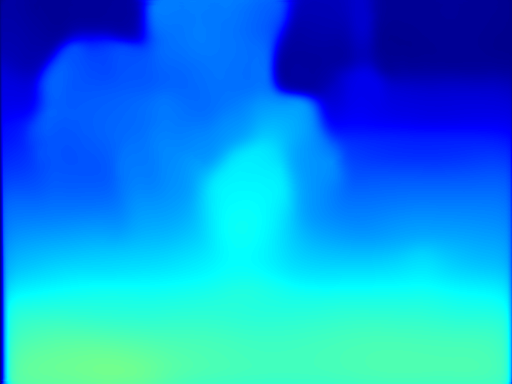}
        \end{tabular}
        \\
        \begin{tabular}{@{}c}
        \includegraphics[width=0.125\linewidth]{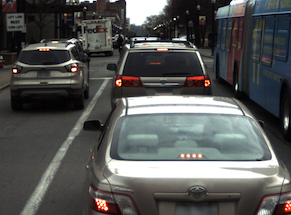}
        \end{tabular}
        &
        \begin{tabular}{@{}c}
        \includegraphics[width=0.125\linewidth]{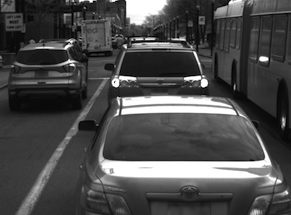}
        \end{tabular}
        &
        \begin{tabular}{@{}c}
        \includegraphics[width=0.125\linewidth]{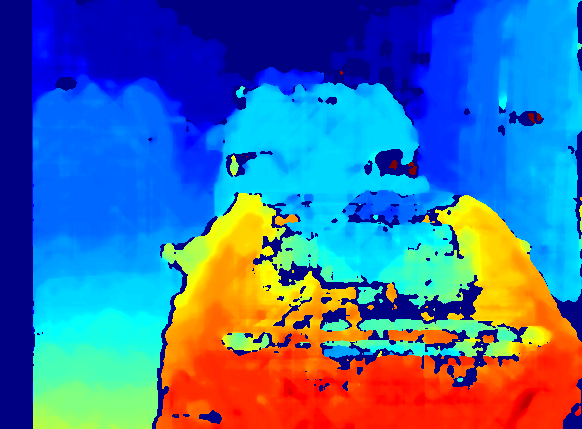}
        \end{tabular}
        &
        \begin{tabular}{@{}c}
        \includegraphics[width=0.125\linewidth]{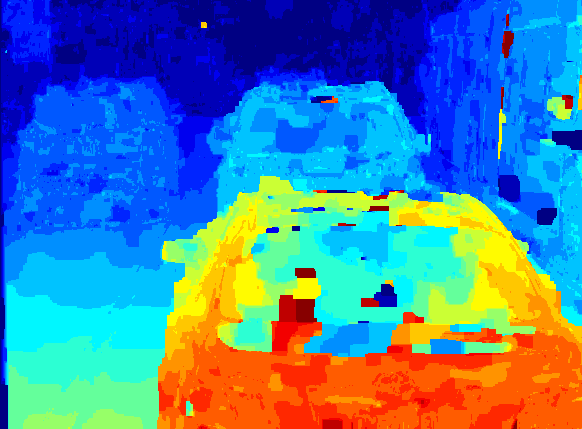}
        \end{tabular}
        &
        \begin{tabular}{@{}c}
        \includegraphics[width=0.125\linewidth]{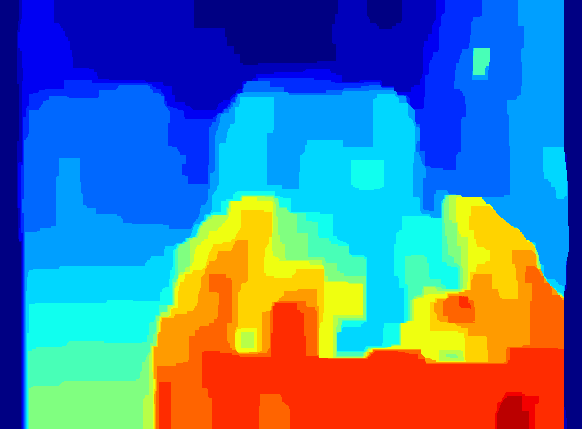}
        \end{tabular}
        &
        \begin{tabular}{@{}c}
        \includegraphics[width=0.125\linewidth]{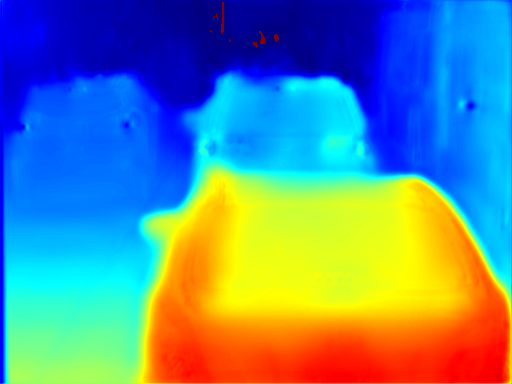}
        \end{tabular}
        &
        \begin{tabular}{@{}c}
        \includegraphics[width=0.125\linewidth]{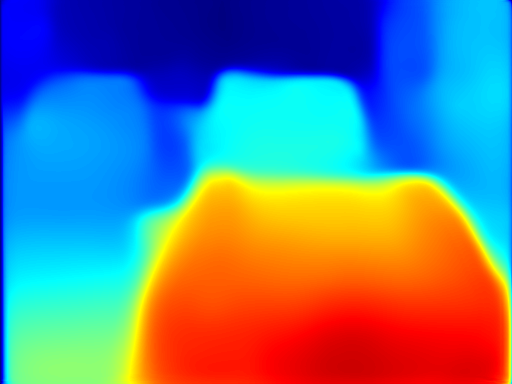}
        \end{tabular}
        \\
        (a) Left VIS
        &
        (b) Right NIR
        &
        (c) CMA
        &
        (d) ANCC
        &
        (e) DASC
        &
        (f) DMC(w.o. seg.)
        &
        (g) Proposed
    \end{tabular}
    \caption{Qualitative results on the evaluation dataset. The results of CMA , ANCC, DASC, DMC(w.o. seg.) are extracted from \cite{zhi2018deep}. The proposed method performs well on the challenging materials like clothes (row 1,2), vegetation (row 2), lights (row 3). And comparing with others, the disparities predicted by the proposed method are cleaner and more reasonable.}
    \label{fig:comparison}
\end{figure*}

\section{Related Work}

\subsection{Unsupervised Depth Estimation}
Garg et al.~\cite{garg2016unsupervised} first proposed to use warping-based view synthesis to learn disparity in an unsupervised way. The right image is first warped to the left view using disparity. Then, the absolute difference between the warped image and the left image, also called reconstruction error or photometric loss, is minimized to supervise disparity predictions. Godard et al.~\cite{godard2017unsupervised} extended this idea by incorporating left-right consistency into the unsupervised loss. Zhou et al.~\cite{zhou2017unsupervised} proposed a framework which simultaneously predicted depth and frame-to-frame relative camera pose, which was trained with photometric loss using consecutive frames from videos. Zhou et al.~\cite{zhouchao2017unsupervised} iteratively train a stereo network by filtering reliable predictions with left-right consistency check. However, unsupervised methods based on photometric loss often fail to predict accurate disparity for cross-spectral images due to the appearance differences. 

\subsection{Cross-spectral Stereo Matching}
A series of robust matching costs were designed for radiometric variations. Mutual
information (MI) measure ~\cite{egnal2000mutual} was extended by incorporating prior probabilities and 2D match surface~\cite{fookes2004multi}. Heo et al.~\cite{heo2011robust} used color formation model explicitly and proposed Adaptive Normalized Cross-Correlation (ANCC) to tackle illumination changes and camera parameter differences. Local self-similarity (LSS)~\cite{torabi2011local} used window-based self similarity descriptor to do dense correspondence measure for thermal-visible videos. Pinggera1 et al.~\cite{pinggera122012cross} showed that dense gradient features based on HOG achieved better performance than MI and LSS descriptors. Aguilera et al.~\cite{aguilera2015lghd} proposed a feature descriptor for matching features points with nonlinear intensity variations. Kim et al.~\cite{kim2015dasc} proposed Dense adaptive self-correlation descriptor (DASC) by improving LSS descriptor with random receptive field pooling. 

Another track of works tried to improve the quality of depth captured by RGBD cameras~\cite{zhu2008fusion,chiu2011improving,de2014improved}. Chiu et al.~\cite{chiu2011improving} proposed a cross-modal adaptation method for cross-spectral stereo matching and fused predictions with depth captured with Kinect. 

For deep learning methods, Aguilera et al.~\cite{aguilera2016learning} learned a similarity measurement of cross-spectral image patches, which is a potential way to learn matching cost for multi-spectrum images. Zhi et al.~\cite{zhi2018deep} utilized deep segmentation maps to improve robustness of cross-spectral stereo matching, while the method required extra semantic annotations and manually designed losses for different materials, which made it hard to apply to other scenes.

\subsection{Image-to-image Translation}
Image-to-image translation converts images from one modality to another, such as style transfer~\cite{gatys2016image,johnson2016perceptual}, colorization~\cite{cheng2015deep}, sketch to image~\cite{zhu2016generative,sangkloy2017scribbler}.
For image translation, training with only L1 loss results in predictions lack of local semantic details. 
Johnson et al.~\cite{johnson2016perceptual} combined per-pixel loss with perceptual loss to train a fast feed-forward network for image transformation. Isola et al.~\cite{isola2017image} proposed a general framework for image-to-image translation using conditional adversarial networks. High-resolution synthetic images can be generated by applying multi-scale structure and novel adversarial loss~\cite{wang2018high}. Later, Zhu et al.~\cite{zhu2017unpaired} proposed CycleGAN to translate image styles with unaligned images from different domains.

We utilize the method of \cite{zhu2017unpaired} to convert images across different spectra, which is a basis of our proposed framework. 

\begin{figure}[t]
\centering
    \begin{tabular}{l|r}
        Step (1) &
        \begin{tabular}{@{}r}
        \includegraphics[height=0.200\linewidth]{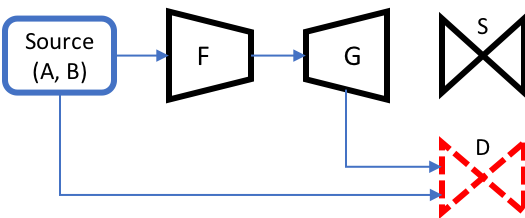}
        \end{tabular}
        \\\hline
        Step (2) &
        \begin{tabular}{@{}r}
        \includegraphics[height=0.200\linewidth]{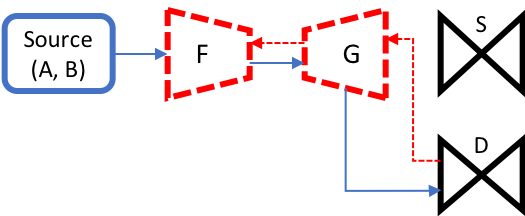}
        \end{tabular}
        \\\hline
        Step (3) &
        \begin{tabular}{@{}r}
        \includegraphics[height=0.200\linewidth]{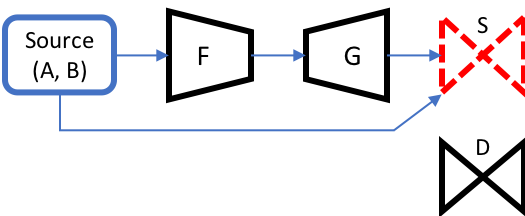}
        \end{tabular}
        \\\hline
        Step (4) &
        \begin{tabular}{@{}r}
        \includegraphics[height=0.200\linewidth]{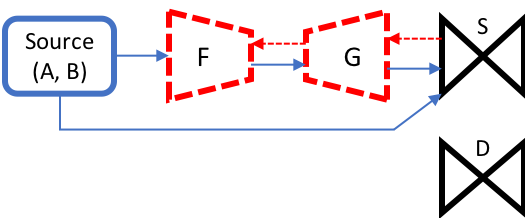}
        \end{tabular}
    \end{tabular}
\caption{The steps of the iterative optimization. Here we omit subscript of spectrum cause the processing of the two modalities is identical. The $F, G, S, D$ represent the $F$ network, the generator network, the stereo matching network, and the adversarial discriminator. The \emph{solid arrows} indicate the directions of data flow during the forward pass, while the \emph{dotted arrows} represent the directions of gradient flow during the backward pass. The \textbf{red dotted} blocks are updated during the corresponding step and the \textbf{black solid} blocks are frozen.}
\label{fig:iterative-optimize}
\end{figure}

\begin{figure*}[htb]
    \centering
    \begin{tabular}{@{}c@{}c@{}c@{}c@{}c}
        \begin{tabular}{@{}c}
        \includegraphics[width=0.185\linewidth]{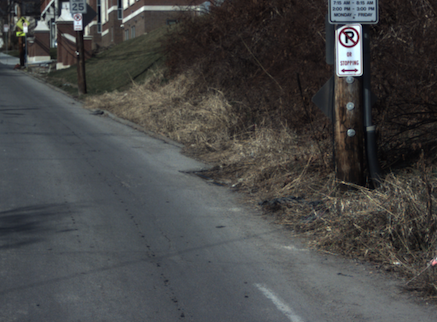}
        \end{tabular}
        &
        \begin{tabular}{@{}c}
        \includegraphics[width=0.185\linewidth]{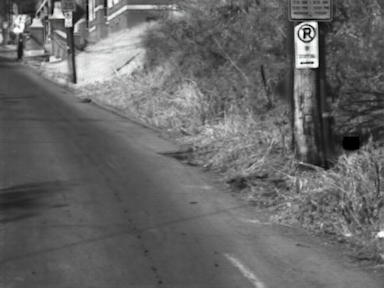}
        \end{tabular}
        &
        \begin{tabular}{@{}c}
        \includegraphics[width=0.185\linewidth]{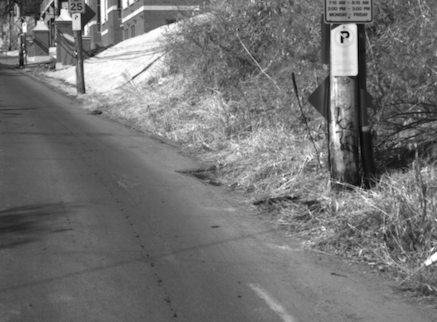}
        \end{tabular}
        &
        \begin{tabular}{@{}c}
        \includegraphics[width=0.185\linewidth]{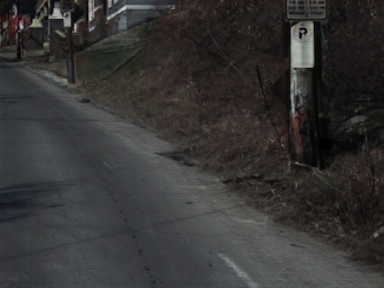}
        \end{tabular}
        &
        \begin{tabular}{@{}c}
        \includegraphics[width=0.185\linewidth]{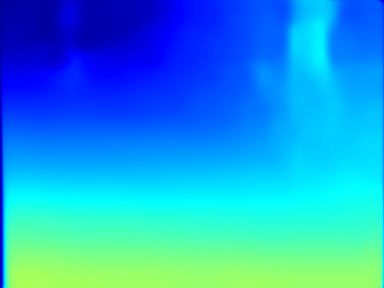}
        \end{tabular}
        \\
        \begin{tabular}{@{}c}
        \includegraphics[width=0.185\linewidth]{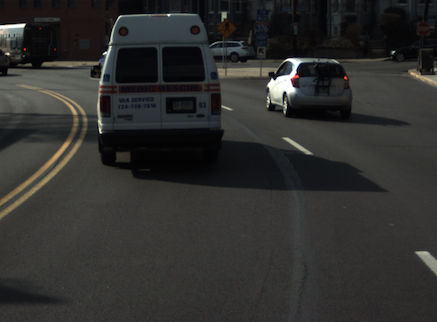}
        \end{tabular}
        &
        \begin{tabular}{@{}c}
        \includegraphics[width=0.185\linewidth]{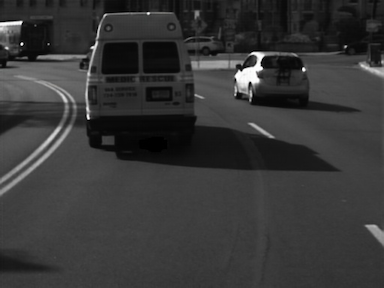}
        \end{tabular}
        &
        \begin{tabular}{@{}c}
        \includegraphics[width=0.185\linewidth]{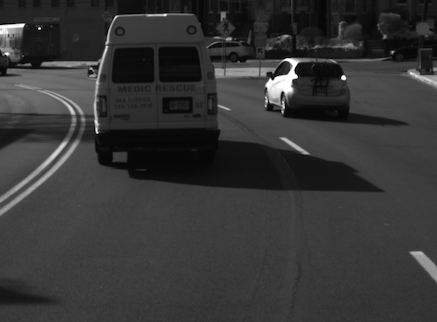}
        \end{tabular}
        &
        \begin{tabular}{@{}c}
        \includegraphics[width=0.185\linewidth]{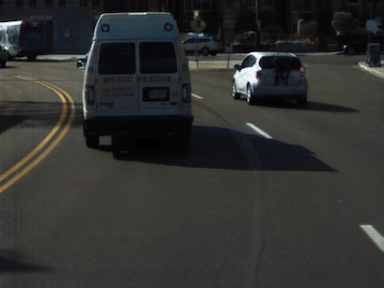}
        \end{tabular}
        &
        \begin{tabular}{@{}c}
        \includegraphics[width=0.185\linewidth]{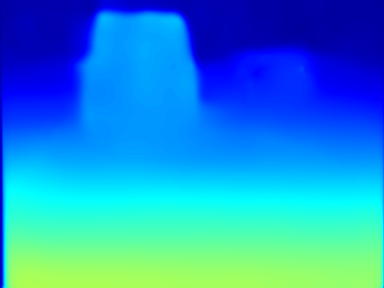}
        \end{tabular}
        \\
        \begin{tabular}{@{}c}
        \includegraphics[width=0.185\linewidth]{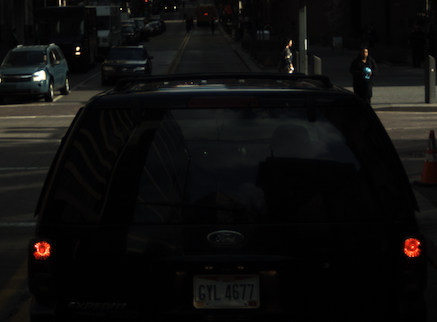}
        \end{tabular}
        &
        \begin{tabular}{@{}c}
        \includegraphics[width=0.185\linewidth]{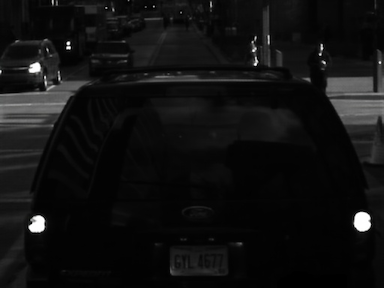}
        \end{tabular}
        &
        \begin{tabular}{@{}c}
        \includegraphics[width=0.185\linewidth]{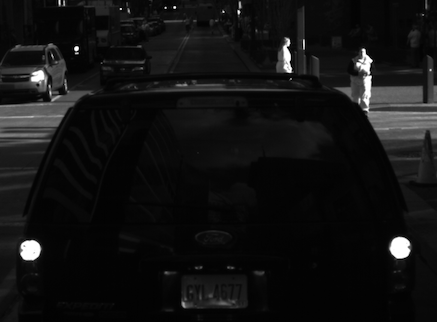}
        \end{tabular}
        &
        \begin{tabular}{@{}c}
        \includegraphics[width=0.185\linewidth]{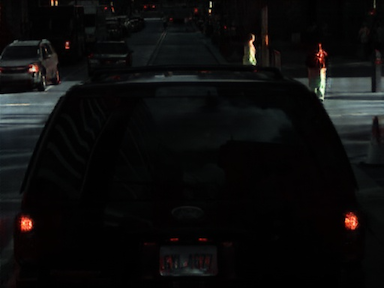}
        \end{tabular}
        &
        \begin{tabular}{@{}c}
        \includegraphics[width=0.185\linewidth]{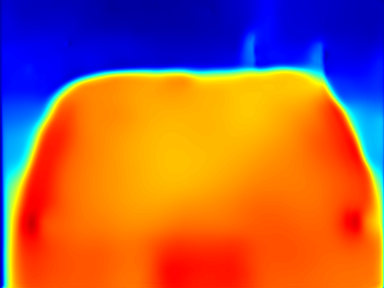}
        \end{tabular}
        \\
        \begin{tabular}{@{}c}
        \includegraphics[width=0.185\linewidth]{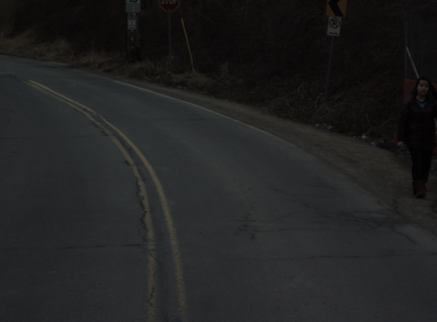}
        \end{tabular}
        &
        \begin{tabular}{@{}c}
        \includegraphics[width=0.185\linewidth]{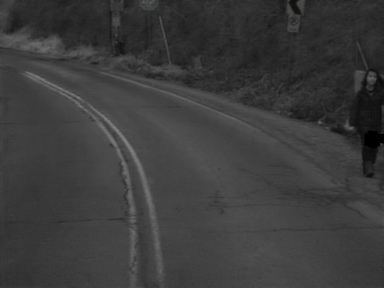}
        \end{tabular}
        &
        \begin{tabular}{@{}c}
        \includegraphics[width=0.185\linewidth]{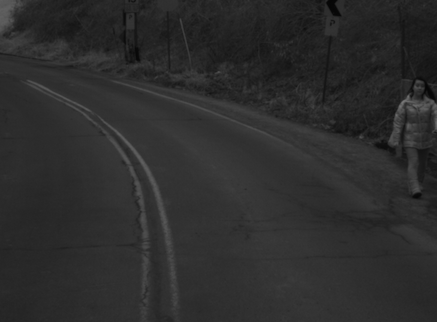}
        \end{tabular}
        &
        \begin{tabular}{@{}c}
        \includegraphics[width=0.185\linewidth]{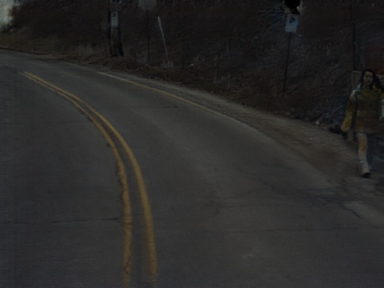}
        \end{tabular}
        &
        \begin{tabular}{@{}c}
        \includegraphics[width=0.185\linewidth]{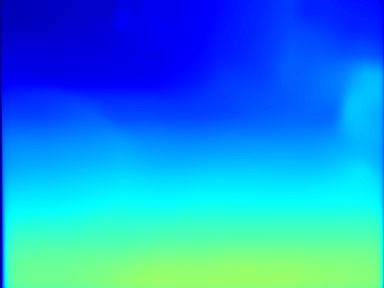}
        \end{tabular}
        \\
        (a) Left VIS
        &
        (b) Left fake NIR
        &
        (c) Right NIR
        &
        (d) Right fake VIS
        &
        (e) Disparity
    \end{tabular}
    \caption{Qualitative results of our proposed approach}
    \label{fig:results}
\end{figure*}
\section{Method}

In this section, we provide a detailed description of each part of the proposed method. Our network can be divided into two parts, the spectral translation network~(STN) and stereo matching network~(SMN). STN is responsible for minimizing the differences between domains, and SMN is responsible for predicting the disparity.

\subsection{Spectral Translation Network}
\label{section: STN}

The goal of the STN is to minimize the appearance variations between different spectra and provide the supervision information to the SMN. To achieve the goal, we proposed a novel style adaptation network F-cycleGAN as STN.

Given any image ${I_A}$ of spectral $A$ and image ${I_B}$ of spectral $B$, we regard ${I_A}$ and ${I_B}$ sampled from two distributions $A$ and $B$.
We can define three mapping functions,
\begin{align}
    F: I_{A,B} \rightarrow X_{A,B} \nonumber, \\
    G_A: X_{A,B} \rightarrow I_A \nonumber, \\
    G_B: X_{A,B} \rightarrow I_B \nonumber,
\end{align}
where $F$ encodes image $I_{A,B}$ to a unified feature space $X$. $G_A$ and $G_B$ are generators which convert features back into images in spectrum $A$ and $B$ respectively. In our implementation, we take the encoder and the decoder of the generator network in CycleGAN~\cite{zhu2017unpaired} as the structure of our $F$ network and $G_{A/B}$ networks.

The network $F$, $G_A$ and $G_B$ are supervised by adversarial losses~\cite{goodfellow2014generative} and cycle-reconstruction loss. The adversarial loss is given by two discriminator networks $D_A$ and $D_B$, which try to differentiate real and fake A or B images. We define $G_B(X_A)$ as $I^{fake}_B$, $G_A(X_B)$ as $I^{fake}_A$, $G_A(X_A)$ as $I_A^{rec}$, and $G_B(X_B)$ as $I_B^{rec}$.

The discriminator $D_A$ aims to distinguish between $I^{fake}_A$ and $I_A$. To train $D_A$, a classification loss $L^{adv,A}_D$ is used to classify $I^{fake}_A$ and $I_A$. The loss for training discriminators is thus defined by

\begin{equation}
L_D^{adv} =  L^{adv,A}_D + L^{adv,B}_D.
\label{equation: adv_D}
\end{equation}

For generator networks, the loss can be mainly divided into two parts, adversarial loss and cycle consistency loss. The adversarial loss aims at fooling the discriminator networks and is given by

\begin{equation}
L_G^{adv} =  L^{adv,A}_G + L^{adv,B}_G,
\label{equation: adv_G}
\end{equation}
where $L^{adv,A}_G$ and $L^{adv,B}_G$ are achieved by maximizing the classification errors of discriminators $D_A$ and $D_B$ (details in ~\cite{zhu2017unpaired}). The cycle consistency loss is,
\begin{align}
\begin{split}
L_G^{cyc} = \frac{1}{N} \sum_{\textbf{p} \in \Omega}& \left\|I_A^{cyc}(\textbf{p})-I_A(\textbf{p})\right\| \\
+ & \left\|I_B^{cyc}(\textbf{p})-I_B(\textbf{p})\right\|,
\label{equation: cyc loss}
\end{split}
\end{align}
where N is the number of pixels, $I_A^{cyc}$ means $G_A(F(I^{fake}_B))$, $I_B^{cyc}$ means $G_B(F(I^{fake}_A))$, and $\Omega$ is the pixel coordinate space. To guarantee the network $F$ maps the images to the same hidden semantic feature space, and prevent the STN from learning disparity, a auxiliary reconstruction loss is introduced to supervise the network:
\begin{align}
\begin{split}
    L_G^{rec} = \frac{1}{N} \sum_{\textbf{p} \in \Omega}& \left\|I_A^{rec}(\textbf{p})-I_A(\textbf{p})\right\| \\
    + &\left\|I_B^{rec}(\textbf{p})-I_B(\textbf{p})\right\|.
\label{equation: rec loss}
\end{split}
\end{align}

Then the final loss for the image transformation network $F,G_A,G_B$ and the adversarial discriminator are given by
\begin{align}
    L_{G} &= \lambda_{c} L_G^{cyc} + \lambda_{r} L_G^{rec} + \lambda_{a} L_G^{adv}
    \label{equation: cycGAN loss}
    \\
    L_{D} &= \lambda_{d} L_D^{adv}
    \label{equation: D loss}
\end{align}

To make the expressions clearer, all the intermediate outputs are summarized as follows,

\begin{equation}
I_A \xrightarrow{F} X_A \xrightarrow{G_B} I^{fake}_{B} \xrightarrow{F} X^{fake}_B \xrightarrow{G_A} I_A^{cyc}
\label{equation: cycle_A}
\end{equation}
\begin{equation}
I_B \xrightarrow{F} X_B \xrightarrow{G_A} I^{fake}_{A} \xrightarrow{F} X^{fake}_A \xrightarrow{G_B} I_B^{cyc}
\label{equation: cycle_B}
\end{equation}
\begin{equation}
X_A \xrightarrow{G_A} I^{rec}_{A} , X_B \xrightarrow{G_B} I^{rec}_{B}
\label{equation: auto-encoder}
\end{equation}

\subsection{Cross-spectral Stereo Matching Network}
\label{section: SMN}
Dispnet~\cite{mayer2016large}, which takes concatenated images as input to directly regress disparities, is adopted as the SMN to predict disparity maps $d^l$,$d^r$ for left and right images. Given rectified cross-spectral image pair $I_{ori}^l$, $I_{ori}^r$, without loss of generality, we assume the spectrum of $I_{ori}^l$ as spectrum $A$ and $I_{ori}^r$ as spectrum $B$. STN is applied to transform the cross-spectral images to the same modality. After that, we concatenate $\{I^l_{ori},G_B(F(I^l_{ori}))\}$ as $I^l$ and $\{G_A(F(I^r_{ori})),I^r_{ori}\}$ as $I^r$ to get the image pair in the same modality, which can be used as the input to the stereo matching network and for the cross-spectral unsupervised loss which will be discussed in the following section.

It should be emphasized that we block the gradients from network inputs back into STN for training stability. It should be noted that the forms of the image used for input and supervision are not required to be identical which will be discussed in benchmark results section.
We apply the training loss from \cite{godard2017unsupervised} which includes appearance matching loss $L_S^{ap}$, disparity smoothness loss ${L_S^{ds}}$, and left-right disparity consistency loss ${L_S^{lr}}$.
We only show the left terms, since the right can be derived similarly.

Based on the left disparity $d^l$, we can get reconstructed left image $\tilde{I^l}$ from $I^r$ with the warping operator $\omega$, which can be described as
\begin{align}
    \tilde{I^l} = \omega(I^r, d^l) \Longleftrightarrow \tilde{I^l}_{x,y}=I^r_{x+d^l_{x,y}, y}.
\end{align}
Since the disparity value might be a float number, $I^r_{x+d^l_{x,y},y}$ is bilinearly sampled at the pixel $(x+d^l_{x,y},y)$. For simplicity, we use a mask to stop calculating the gradients for the pixels which are unable to be warped (e.g. pixels out of bound).

The appearance matching loss $L_S^{ap}$ encourages the reconstructed image to appear similar to the original image by comparing structure and intensity. We let $\delta(I_1, I_2)$ be the structural similarity function~\cite{wang2004image} and $\tilde{I^l}$ be the reconstruction of $I^l$ from $I^r$. The appearance matching loss can be described as

\begin{align}
\begin{split}
L_S^{ap,l} = \frac{1}{N} \sum_{\textbf{p} \in \Omega} &\alpha \frac{1 - \delta(I^l, \tilde{I}^l)(\textbf{p})}{2} + \\ &(1-\alpha)\left \| I^l(\textbf{p}) - \tilde{I}^l(\textbf{p}) \right \|,
\label{eq:loss ap}
\end{split}
\end{align}
where $\alpha$ denotes the weight coefficient for the structural dissimilarity function and L1 reconstruction loss. The loss $L_S^{ds}$ enforces the disparity smoothness,

\begin{align}
    \begin{split}
        L_S^{ds,l} = \frac{1}{N} \sum_{\textbf{p} \in \Omega} (\left | \partial_x d^l   \right | e^{-\left \| \partial_x I^l \right \|} + \left | \partial_y d^l   \right | e^{-\left \| \partial_y I^l \right \|})(\textbf{p}),
\label{eq:cds}
    \end{split}
\end{align}
where $\partial d^l$ and $\partial I^l$ means the gradients of $d^l$ and $I^l$. The loss $L_S^{lr,l}$ regularizes the consistency of the left disparity and the right disparity,

\begin{align}
    L_S^{lr,l} = \frac{1}{N} \sum_{\textbf{p} \in \Omega}|d^l(\textbf{p})-\omega(d^r, d^l)(\textbf{p})|.
\end{align}
Then the final loss for the SMN network is given by
\begin{align}
    \begin{split}
            L_{SMN} &= \alpha_{ap} (L_S^{ap,l} + L_S^{ap,r}) + \alpha_{ds} (L_S^{ds,l} + L_S^{ds,r})\\ &+ \alpha_{lr} (L_S^{lr,l} + L_S^{lr,r}).
    \label{equation: SMN loss}
    \end{split}
\end{align}

To further improve the performance, we introduce an auxiliary loss for the STN. First we can get the warped original images $\tilde{I^l_{ori}}=\omega(I^r_{ori}, d^l)$ and $\tilde{I^r_{ori}}=\omega(I^l_{ori}, d^r)$ with disparity prediction, then the auxiliary loss $L_G^{aux}$ is defined by

\begin{align}
\begin{split}
L_G^{aux} = \alpha_{aux} \frac{1}{N} \sum_{\textbf{p} \in \Omega}& \left\|G_B(F(I^l_{ori}))(\textbf{p})-\tilde{I^l_{ori}}(\textbf{p})\right\| \\
+ & \left\|G_A(F(I^r_{ori}))(\textbf{p})-\tilde{I^r_{ori}}(\textbf{p})\right\|,
\label{equation: aux loss}
\end{split}
\end{align}
which attempts to tackle appearance variations and enhance the robustness of STN. There is a possibility that the reconstruction may encode both the disparity and spectral differences. We hold that by the cycle loss and reconstruction loss in Equ. \ref{equation: cyc loss} and Equ. \ref{equation: rec loss}, we can prevent the STN from learning disparity. 

\subsection{Iterative Optimization}
\label{section: iterative optimize}

We will introduce our iterative optimization approach in this section. All the losses required are presented in the Equ. \ref{equation: cycGAN loss}, Equ. \ref{equation: D loss}, Equ. \ref{equation: SMN loss}, and Equ. \ref{equation: aux loss}.
For simplicity, we omit subscript of spectrum for $G_A,G_B,D_A,D_B$ because the optimization for the two modalities is identical.

Figure \ref{fig:iterative-optimize} shows the gradient flow across different network blocks. A randomly sampled cross-spectral image pair is provided to the entire system in each iteration. For the step (1), we train the $D$ network by loss $L_D$ from Equ. \ref{equation: D loss}, which encourage the discriminator to distinguish between real and fake images. Then for the step (2), we train the $F$ network and $G$ network by loss $L_G$ from Equ. \ref{equation: cycGAN loss}. The stereo network $S$ is trained in step (3) with the loss $L_{SMN}$ from Equ. \ref{equation: SMN loss} by taking the translation results from $G$ network as supervision. Finally, we use loss $L_{aux}$ from Equ. \ref{equation: aux loss} to train the $F$ network and $G$ network again for global optimization. The whole framework is first trained with several warmup epochs, using only step (1) and step (2), during which the stereo matching network is not trained. After the warmup stage, all four steps are used for further training.  

\begin{table*}[t]
  \begin{center}
    \caption{Quantitative results. The RMSE of disparity for each material is evaluated. The RMSE results and execute times of CMA, ANCC, DASC, DMC(w.o. seg.), DMC(w. seg.) are extracted from \cite{zhi2018deep}, where the DMC(w. seg.) means the method of \cite{zhi2018deep} with material-aware confidence. The proposed methods are tested on a single NVIDIA TITAN Xp GPU, which is the same as \cite{zhi2018deep}. The network structure changes (row 7-10) lead to the improvement of performance.}
    \label{table: RMSE}
	\begin{tabular}{c|c|c|c|c|c|c|c|c|c|c} 
    \hline
    Method&Common&Light&Glass&Glossy&Veg.&Skin&Clothing&Bag&Mean&Time(s)\\
    \hline
    CMA &1.60&5.17&2.55&3.86&4.42&3.39&6.42&4.63&4.00&227\\
    ANCC &1.36&2.43&2.27&2.41&4.82&2.32&2.85&2.57&2.63&119 \\
    DASC &0.82&1.24&1.50&1.82&1.09&1.59&\textbf{0.80}&1.33&1.28&44.7 \\
    DMC(w.o. seg.) &\textbf{0.51}&1.08&1.05&1.57&0.69&\textbf{1.01}&1.22&0.90&1.00&0.02 \\
    \hline
    DMC(w.seg.) &0.53&0.69&0.65&0.70&0.72&1.15&1.15&0.80&0.80&0.02 \\
    \hline
    Only SMN&1.25&1.37&1.13&1.65&1.07&1.50&1.18&0.96&1.27&0.02\\
    STN + SMN &1.13&1.55&1.05&1.52&0.89&1.23&1.14&0.98&1.18&0.04\\
    STN(F) + SMN&1.24&1.02&0.92&1.32&0.79&1.10&1.03&0.92&1.04&0.04 \\
    STN(F) + SMN(aux)(ori) &0.75&0.86&\textbf{0.63}&1.05&0.81&1.16&0.99&\textbf{0.74}&0.87&0.02 \\
    STN(F) + SMN(aux)&0.68&\textbf{0.80}&0.67&\textbf{1.05}&\textbf{0.68}&1.04&0.98&0.80&\textbf{0.84}&0.04 \\
    \hline
    Full Method&0.68&\textbf{0.80}&0.67&\textbf{1.05}&\textbf{0.68}&1.04&0.98&0.80&\textbf{0.84}&0.04 \\
    \hline
    \end{tabular}
  \end{center}
\end{table*}
\section{Experiments}

In this section, an evaluation of our method is performed on the benchmark dataset, and detailed analysis is given.

The network is trained on rectified cross-spectral stereo image pairs without any supervision in the form of ground truth disparity or depth. We evaluate on the PittsStereo-RGBNIR dataset proposed by \cite{zhi2018deep} which covers many material categories including lights, glass, glossy surfaces, vegetation, skin, clothing and bags. This dataset was captured by a visible (VIS) and near infrared (NIR) camera pairs. We define the left VIS as spectrum $A$ and right NIR as spectrum $B$. The Left VIS consists of three spectral bands while the right NIR consists of only one band. For the simplicity of implementation, we convert NIR images into three channels.

\subsection{Implementation Details}
\subsubsection{Architecture}
The $G$ network and $D$ network followed \cite{zhu2017unpaired} which has shown impressive results for image-to-image translation. The $F$ network contains 4 residual blocks \cite{he2016deep} and two stride-2 convolutions for down-sampling which is similar to the $G$ network.

We used the DispNet~\cite{mayer2016large} as our stereo matching network SMN, and for the training stability, multi-scale predictions of SMN are applied following~\cite{godard2017unsupervised}. The weights of the STN were initialized from a Gaussian distribution with zero mean and 0.02 standard deviation, and the weights of the SMN were initialized with Kaiming initialization~\cite{he2015delving}.

\subsubsection{Parameters}
The SMN predicts the disparity directly instead of the ratio between disparity and image width. The disparity predictions are clamped to the range of zero to the image width. A scaling factor $\eta = 0.008$ is multiplied to the predictions for stable optimization.

The weights of the losses in STN are set to $\lambda_{c} = 10$, $\lambda_{r} = 5$, $\lambda_{a} = 1$, $\lambda_{d} = 1$, and the weights of losses in SMN are $\alpha_{ap}=1$, $\alpha_{ds}$ = 0.2, $\alpha_{lr}$ = 0.1, $\alpha_{aux} = 20$. We use $5\times5$ window for  calculating the structural similarity $\delta$, and the $\alpha$ in Equ. \ref{eq:loss ap} is set to $0.9$.

\subsubsection{Training and Testing}
The entire network contains about 54 million trainable parameters, of which 33 million parameters are in SMN.
The dataset is split into two sets for training (40000 pairs) and testing (2000 pairs), which is the same as \cite{zhi2018deep}. The STN and SMN are trained on 40000 cross-spectral image pairs with Adam optimizer~\cite{kingma2014adam} (batch size = 16 and learning rate = 0.0002).

For data augmentation, we flip the input images of STN horizontally with a 50\% chance. Input images are resized into $512{\times}384$ for the entire network. We perform an instance normalization on the images provided to the SMN as input. The training process takes about 34 hours using 8 Nvidia TITAN Xp GPUs. The network is first trained with 15 warm-up epochs (with only step 1, 2, the SMN is not trained during this stage), and then trained with all 4 steps for 10 epochs.

For testing, the predicted disparity maps are bilinearly upsampled to the original size with disparity values multiplied with the horizontal scaling factor. 5030 sparse points on 2000 testing images are evaluated to compute the root mean square error.

\subsection{Benchmark Results}
\label{benchmark}

For the sake of comparison, we choose the root mean square error (RMSE) as an indicator for our comparison, and we calculate the RMSE of each material category and obtain the average value as the final result \textbf{Mean}, following \cite{zhi2018deep}. We have tested five network structure choices: \textit{only SMN}, \textit{STN+SMN}, \textit{STN(F)+SMN}, \textit{STN(F)+SMN(aux)(ori)}, and \textit{STN(F)+SMN(aux)}. 

For \textit{only SMN}, STN is not employed and the cross-spectral image pairs are directly used as the unsupervised supervision. \textit{STN+SMN} employs the original Cycle-GAN as the spectral translation network. For the \textit{STN(F)} series, we use our proposed F-cycleGAN as the STN. \textit{(aux)} represents using the auxiliary loss during training. The \textit{STN(F) + SMN(aux)(ori)} means the original image pairs instead of concatenated image pairs are used as inputs. All the methods with \textit{STN} except \textit{STN(F) + SMN(aux)(ori)} take concatenated original images and translated fake image pairs from STN as the inputs of SMN. We found that using only the NIR image and the fake NIR image in the unsupervised loss of SMN achieved better results, thus in all of our experiments, we employ only NIR images for the unsupervised supervision of SMN. 

We compare the performance of the proposed method with other cross-spectral stereo matching methods like CMA\cite{chiu2011improving}, ANCC\cite{heo2011robust}, DASC\cite{kim2015dasc}, and DMC\cite{zhi2018deep}. Table \ref{table: RMSE} presents the comparison with disparity RMSE and execution time. For DMC\cite{zhi2018deep}, they incorporate material-aware confidence into the disparity prediction network, which requires semantic segmentation labels and manually defined loss for each kind of material. For fair comparison, we also list their results without the material-aware confidence. 

On average, our approach outperforms other methods without extra human intervention. On lights, glass, glossy surface, and bag, our approach performs better than others. Table \ref{table: RMSE} also presents the changes in the results of our three comparative experiments, \textit{STN+SMN}, \textit{STN(F)+SMN}, and \textit{STN(F)+SMN(aux)}. The results show that the F-cycleGAN and the framework for jointly training are able to improve the performance of unsupervised stereo matching. We also find that it is still hard to translate the appearance of clothing between VIS and NIR by the STN, possibly because the material of clothing is more variable than others, which leads to an unstable correspondence. 

\subsection{Visualization Results}

Figure \ref{fig:results} presents the visualized results of the proposed method which suggests that the proposed approach is able to handle the illumination variations between different spectra. Comparing to other unsupervised methods in Figure \ref{fig:comparison}, our method provides cleaner and more reasonable disparity predictions.

\section{Conclusion}

We have presented an unsupervised cross-spectral stereo matching method which can be trained in an end-to-end way without extra data or excessive human intervention. We propose F-cycleGAN based on the work of the \cite{zhu2017unpaired} as STN, which is able to minimize the appearance variations between different spectra without the loss of geometric information and improve the robustness of the stereo matching network SMN. Our experimental results show that our method outperforms other state-of-the-art methods. Our approach can be directly applied to other spectra, such as short-wave infrared or medium-wave infrared images.

In the future, we expect to further enhance the capabilities of the STN network for subtle visual differences. The structural similarity loss in the unsupervised loss of SMN, which is illumination sensitive, could also be improved to better supervise the stereo matching network.

\fontsize{9.0pt}{10.0pt} \selectfont
\bibliography{reference}
\bibliographystyle{aaai}

\end{document}